# FairMed-XGB: A Bayesian-optimised multi-metric framework with explainability for demographic equity in emergency medicine

Mitul Goswami [a], Romit Chatterjee [a,*], Arif Ahmed Sekh [b]

[a] School of Computer Engineering, KIIT Deemed to be University, Patia, Bhubaneswar, 751024, India
[b] Bio-AI Lab, UiT the Arctic University of Norway, Tromso, Norway

ABSTRACT

*Background:* Machine learning models in emergency medicine often exhibit demographic biases, particularly gender disparities, which can compromise fairness, clinical trust, and decision-making reliability. Existing mitigation approaches typically optimise isolated fairness metrics, failing to address the multi-dimensional nature of bias while maintaining predictive performance and interpretability.
*Methods:* We propose FairMed-XGB, a Bayesian-optimised, multi-metric fairness framework integrated into an XGBoost classifier. The model incorporates a composite fairness-aware loss function combining Statistical Parity Difference, Theil Index, and Wasserstein Distance. Bayesian optimisation is used to balance fairness and predictive accuracy. The framework is evaluated on seven cohorts derived from the MIMIC-IV-ED and eICU datasets. SHAP-based explainability is employed to analyse feature contributions and identify bias-related patterns.
*Results:* FairMed-XGB significantly reduces gender-based disparities across all cohorts. Statistical Parity Difference decreases by 40–51% on MIMIC-IV-ED and 10–19% on eICU datasets. Theil Index values reduce by four to five orders of magnitude, approaching near-zero levels, while Wasserstein Distance decreases by 20–72%. These improvements are achieved with minimal impact on predictive performance (AUC-ROC reduction <0.02). SHAP analysis demonstrates reduced dependence on gender-proxy features and improved distributional fairness.
*Conclusions:* FairMed-XGB provides a robust and interpretable framework for mitigating demographic bias in emergency medicine prediction models. By jointly optimising multiple fairness objectives and preserving model performance, it enables equitable and trustworthy AI deployment in high-stakes clinical settings. The integration of explainability further supports transparency and clinical adoption.

## 1. Introduction

Machine learning (ML) has emerged as a transformative force in healthcare, particularly in high-stakes emergency medicine settings such as intensive care units (ICUs) and emergency departments. Advanced predictive models are increasingly deployed to forecast patient outcomes, optimise resource allocation, and guide treatment protocols, leveraging vast datasets to enhance clinical decision-making [1,2]. For instance, algorithms trained on electronic health records (EHRs) can predict sepsis onset, mortality risk, or likelihood of readmission with remarkable accuracy, enabling proactive interventions that save lives and reduce costs [3]. These models hold immense potential to streamline workflows, minimise human error, and improve equity in care delivery. However, this growing reliance on data-driven systems underscores a pressing ethical challenge: biases embedded in training data or algorithmic design can perpetuate or exacerbate disparities among vulnerable populations. Studies reveal that demographic imbalances, such as the underrepresentation of female patients in cardiovascular datasets or racial disparities in pain assessment algorithms, often lead to skewed predictions, misdiagnoses, or inequitable resource distribution [4,5]. In emergency medicine, where timely and unbiased decisions are paramount, such flaws risk deepening existing healthcare inequities, disproportionately affecting marginalised groups. Thus, ensuring fairness in ML-driven clinical tools is not merely a technical concern but a moral imperative to uphold equity and trust in modern healthcare systems.

The efficacy of machine learning models in healthcare hinges on the representativeness and quality of training data; yet, pervasive demographic biases in clinical datasets often undermine their fairness. Biases arise from systemic inequities in data collection, such as the underrepresentation of minority groups, imbalanced outcomes across demographics, or incomplete documentation of socioeconomic factors [6,7]. For instance, models trained on ICU datasets with skewed gender ratios may systematically misestimate mortality risks for female patients, while algorithms leveraging racially homogeneous data might inaccurately triage patients from underrepresented backgrounds [8]. These biases propagate through the predictive pipeline, amplifying disparities in critical outcomes. A 2022 study of MIMIC-IV data revealed that models predicting sepsis exhibited significantly lower recall for Black patients compared to White patients, exacerbating delays in life-saving interventions [9]. Similarly, socio-economic biases embedded in triage algorithms have been shown to disproportionately divert resources away from low-income populations, compounding existing access barriers [10]. Such failures not only erode trust in clinical AI but also risk perpetuating cycles of inequity, where marginalised groups face compounded disadvantages in acute care settings. Addressing these challenges requires frameworks that explicitly identify and mitigate data-driven biases while preserving model utility—a gap this work seeks to bridge.

* Corresponding author.
*E-mail address:* chatterjeeromit86@gmail.com (R. Chatterjee).

Existing bias mitigation techniques, such as reweighting and adversarial debiasing, often prioritise isolated fairness metrics, such as statistical parity, while neglecting the multifaceted nature of demographic equity in healthcare [11,12]. For example, reweighting methods may balance dataset representation across groups but fail to address disparities in model performance metrics such as false-negative rates, which are critical in life-threatening conditions. Similarly, adversarial approaches that suppress sensitive attribute correlations risk degrading predictive accuracy or ignoring intersectional biases arising from overlapping demographic factors (e.g., race and gender) [13]. Furthermore, current frameworks seldom harmonise competing fairness objectives, such as equalized odds (equality in error rates) and distributive fairness (alignment of prediction distributions across groups), leading to suboptimal trade-offs in clinical settings [14]. Compounding this issue is the lack of explainability in fairness-aware models: many methods operate as "black boxes," offering clinicians limited insight into how bias mitigation impacts decision logic or feature importance [15]. This opacity hinders trust and adoption, as healthcare providers require transparent tools to validate model behaviour against ethical and clinical standards.

This work introduces FairMed-XGB, a novel framework designed to advance demographic equity in critical healthcare prediction models through three synergistic innovations. First, it employs Bayesian-optimised multi-metric fairness, dynamically balancing equalized odds, Theil index, and Wasserstein distance to harmonise accuracy with subgroup equity, recognising that no single fairness metric can fully characterise demographic equity in complex healthcare environments. The selected metrics were chosen to capture three distinct dimensions of fairness: demographic parity (SPD), distributional inequality (Theil Index), and probability distribution alignment (Wasserstein Distance), thereby providing a more holistic assessment than any individual fairness measure alone. Unlike prior methods that fix fairness constraints statically, this approach adaptively tunes hyperparameters to optimise trade-offs across heterogeneous ICU datasets, such as eICU and MIMIC-IV ED. Second, the framework prioritises demographic equity by explicitly mitigating gender-based disparities in clinical outcomes, a critical gap in existing emergency medicine models, through rigorous evaluation of prediction distributions and error rates across sub-groups. Third, FairMed – XGB integrates SHAP to provide granular, clinician-actionable insights into feature contributions, bridging the explainability gap in fairness-aware ML. By linking SHAP values to bias, the framework ensures transparency in both model decisions and mitigation strategies. Furthermore, unlike many existing healthcare fairness studies that evaluate only against unconstrained baseline models, FairMed-XGB is benchmarked against established fairness mitigation strategies, enabling a more rigorous assessment of its effectiveness and practical value.

## 2. Related studies

### 2.1. Medical tabular data for machine and deep learning

Medical tabular datasets, such as MIMIC-IV ED [16] and eICU [17], are pivotal for training ML models in emergency medicine. These datasets capture structured EHRs, including demographics, vitals, and lab results, enabling predictive tasks like sepsis detection. However, inherent biases, such as underrepresentation of minority groups or imbalanced outcomes, compromise model fairness. For instance, MIMIC-IV ED exhibits gender disparities in cardiovascular entries, while eICU data reflects racial imbalances in sepsis prediction recall. Recent studies highlight the prevalence of socioeconomic and racial bias variables in such datasets, necessitating preprocessing or algorithmic mitigation. Table 1 summarises common datasets and associated bias variables.

### 2.2. Bias detection methods in medical data

Detecting biases in medical datasets requires systematic methodologies to identify and quantify inequities. Disparate Impact Analysis [23] compares favorable outcome ratios across privileged and marginalised groups (e.g., ICU admission rates by race) to flag systemic disparities. Causal Graph Analysis [24] uses directed acyclic graphs (DAGs) to uncover confounding variables (e.g., socioeconomic status) and isolate bias pathways. Adversarial Debiasing [25] trains models with adversarial networks to minimise bias during learning, though at times it may reduce predictive accuracy. Fairness Audits [26] assess subgroup performance using metrics like False Positive Rate (FPR) and False Negative Rate (FNR) to reveal disparities (e.g., in sepsis predictions for Black patients). Counterfactual Testing [27] alters sensitive attributes (e.g., gender) to check prediction invariance, exposing causal biases. Reweighting Techniques [28] adjust data weights (e.g., oversampling female cardiovascular cases) to address representation imbalance. Lastly, Subgroup Performance Analysis [29] evaluates models across demographic splits (e.g., age, ethnicity), as shown in MIMIC-IV studies [30].

Recent studies have also highlighted the importance of identifying demographic bias at the dataset level prior to model development. Distribution-based approaches, including Earth Mover's Distance (EMD) and related distributional divergence measures, compare feature and class distributions across demographic groups to identify representation imbalances before model training. Such methods provide valuable insights into whether fairness violations originate from underlying data characteristics or from the learning algorithm itself. These approaches complement model-level fairness assessment by enabling early detection of demographic disparities in healthcare datasets.

**Table 1**
Bias variables in medical tabular datasets.

| Dataset | Description | Bias Variables |
|---|---|---|
| MIMIC-IV [16] | ICU data from Beth Israel Hospital | Gender, Race, Age |
| eICU [17] | Multi-center ICU data | Race, Socioeconomic status |
| NHANES [18] | National health survey data | Ethnicity, Income, Education |
| UK Biobank [19] | Large-scale UK biomedical database | Ethnicity, Socioeconomic status |
| All of Us [20] | NIH-funded diverse cohort data | Geographic diversity, Race |
| Framingham Heart [21] | Longitudinal cardiovascular study | Historical cohort |
| BRFSS [22] | CDC behavioral risk factor survey | Self-reporting bias, Income |

### 2.3. Bias mitigation methods in medical data

Bias mitigation in medical data employs techniques to reduce disparities in model outcomes across demographic groups. Reweighting [31] adjusts sample weights to balance underrepresented cohorts, such as oversampling female patients in cardiovascular datasets. Adversarial debiasing [32] trains models with adversarial networks to minimise correlations between predictions and sensitive attributes (e.g., race in sepsis risk models). Fairness constraints [33] integrate fairness metrics (e.g., equalized odds) into the loss function, penalising discriminatory predictions during training. Data augmentation [34], including synthetic minority oversampling (SMOTE), addresses class imbalance in rare disease datasets. Causal mitigation methods [35] leverage directed acyclic graphs to adjust for confounders like socioeconomic status, ensuring unbiased causal relationships in outcomes. Post-processing techniques [36] recalibrate decision thresholds (e.g., ICU discharge predictions) to align error rates across subgroups. Fair meta-learning [37] adapts models to heterogeneous populations by transferring fairness-aware knowledge across datasets. While these methods reduce bias, challenges persist in preserving clinical accuracy and addressing intersectional disparities (e.g., race-gender interactions).

Recent work has specifically addressed gender bias evaluation and mitigation in disease prediction models. Notably, a 2026 study [38] proposed a systematic framework for evaluating and mitigating gender bias in clinical prediction models for chronic diseases, demonstrating that standard pre-processing debiasing techniques alone are insufficient when gender interacts with disease-specific clinical features. The study employed multiple fairness metrics including Equalized Odds and Calibration Error alongside SPD, and showed that model-level fairness interventions combined with feature-level analysis can achieve more comprehensive bias reduction. This finding aligns with the multi-metric approach adopted in the present study and supports the use of SHAP-based feature analysis to identify gender-proxy variables. Similarly, Uddin et al. [39] proposed a fairness-preserving framework that integrates dataset-level bias quantification using Earth Mover's Distance (EMD) with model-level fairness evaluation. Their study evaluated five supervised learning algorithms across multiple fairness definitions and healthcare datasets, highlighting the importance of jointly considering underlying data bias and model-level fairness. These findings further reinforce the need for comprehensive fairness evaluation that considers both dataset characteristics and predictive model behaviour.

### 2.4. Explainable bias detection and mitigation methods

Explainable methods enhance transparency in identifying and addressing biases in medical AI. SHAP (SHapley Additive exPlanations) [40] quantifies feature contributions to predictions, exposing biases in variables like race or insurance type. LIME (Local Interpretable Model-agnostic Explanations) [41] generates local explanations to highlight biased decision logic (e.g., triage disparities in eICU data). FairML [42] audits models via input perturbation to trace biases to specific features (e.g., ZIP code proxies for race). Interpretable fairness constraints [43], such as equalized odds penalties, integrate fairness metrics into model training while providing decision-boundary transparency. Counterfactual explanations [44] test prediction invariance under hypothetical changes to sensitive attributes (e.g., altering gender in mortality risk models), isolating causal bias pathways. Rule-based models [45], like fairness-aware decision trees, enforce interpretable fairness rules (e.g., equitable ICU discharge criteria).

## 3. Methodology

This section details the FairMed-XGB framework, a systematic approach designed to detect and mitigate gender-based prediction bias in clinical machine learning models while preserving predictive performance and explainability. The methodology comprises four core stages: (1) data preprocessing, (2) pre-mitigation bias detection and analysis, (3) the formulation and integration of a fairness-aware loss function, and (4) post-mitigation training and evaluation. The mathematical formulation of the framework is presented in Algorithm 1.

**Algorithm 1**: FairMed–XGB: Bayesian Optimised Multi-Metric Fairness Learning

**INPUT:**
Clinical dataset
$D = \{(x_i, y_i, a_i)\}_{i=1}^{n}$
where
$x_i \in \mathbb{R}^d$: feature vector
$y_i \in \{0, 1\}$: binary clinical label
$a_i \in \{0, 1\}$: sensitive attribute (gender)
**OUTPUT:**
Fairness-aware model $f^*$
**Data Preprocessing**
Encode categorical features:
$x_{i,j} \leftarrow \varphi(x_{i,j})$
Standardise continuous features:
$x_{i,j} \leftarrow (x_{i,j} - \mu_j)/\sigma_j$
Split $D$ into $D_{Train}$ and $D_{Test}$ (stratified on $a_i$)
**Baseline Training**
Initialize model:
$f(x) = 0$
*For* $t = 1$ to $T$ *do*
Compute predictions:
$\widehat{y}_i = \sigma(f(x_i)) = \frac{1}{(1 + e^{-f(x_i)})}$
Compute gradient:
$g_i = \frac{\delta L_{Pred}}{\delta(f(x_i))}$
Fit regression tree $h_t(x)$ to $\{g_i\}$
Update model:
$f(x) \leftarrow f(x) + \eta_t h(x)$
*End For*
**Bias Quantification**
Define groups:
$G_0 = \{i \mid a_i = 0\}$
$G_1 = \{i \mid a_i = 1\}$
Compute Statistical Parity Difference:
$SPD = P(\widehat{Y} = 1 \mid A = 1) - P(\widehat{Y} = 1 \mid A = 0)$
Compute Theil Index:
$Theil = \sum_{i=1}^{m} \frac{\widehat{p}_i}{p} \ln\left(\frac{\widehat{p}_i}{p}\right), where\ \ddot{p} = \frac{1}{m} \sum_{i=1}^{m} \widehat{p}_i$
Compute Wasserstein Distance:
$W = \int_0^1 |F_1(t) - F_0(t)| dt$
**Fairness Aware Objective**
Define fairness loss:
$\mathscr{L}_{fair} = w_1 \cdot SPD + w_2 \cdot Theil + w_3 \cdot W$
Define total loss:
$\mathscr{L}_{total} = \mathscr{L}_{log} + \lambda \cdot (\mathscr{L}_{fair})$
**Bayesian Optimisation**
Initialize search space:
$\theta = (\lambda, w_1, w_2, w_3)$
*For* $k = 1$ *to* $K$ *do*
Train model minimizing $\mathscr{L}_{total}$
Evaluate validation objective:
$J(\theta) = \alpha * AUC - (1 - \alpha) * \mathscr{L}_{fair}$
Update $\theta$ using Gaussian Process surrogate
*End For*
Select optimal $\theta^*$
**Final Training**
Retrain model on $D_{train}$ using $\theta^*$
Obtain final model $f^*$
**Explainability**
*For* each feature*j do*
Compute *SHAP* value $\varphi_j$
Compute group disparity:
$\Delta\varphi_j = \mathrm{E}[\varphi_j \mid a = 0] - \mathrm{E}[\varphi_j \mid a = 1]$
*End For*
***return f****

### 3.1. Data preprocessing

We utilise two publicly available critical-care datasets: MIMIC-IV-ED

[16] and the eICU Collaborative Research Database [17]. To ensure consistency and focus, gender is treated as a binary sensitive attribute $A \in \{0, 1\}$, where $A = 1$ denotes *Male* and $A = 0$ denotes *Female*. While gender was modelled as a binary sensitive attribute in this study due to the structure of the MIMIC-IV-ED and eICU databases, we acknowledge that gender exists on a broader spectrum and cannot be fully represented by binary categories alone. Records with undefined or non-binary gender labels were not available in sufficient numbers within the selected cohorts to support statistically meaningful subgroup analysis. The proposed FairMed-XGB framework is not inherently restricted to binary sensitive attributes and can be extended to multi-class demographic settings by computing fairness metrics across multiple groups and incorporating aggregated disparity penalties into the optimisation objective. Future work will evaluate the framework in more inclusive healthcare datasets containing non-binary and gender-diverse populations.

All records with undefined or non-binary gender values are excluded from the analysis. Categorical features are encoded using label encoding, and continuous features are normalised. Each dataset is partitioned into training and testing subsets using an 80/20 stratified split, preserving the original distribution of the sensitive attribute and target classes. Formally, let $X \in \mathbb{R}^{m \times d}$ represent the feature matrix, $Y \in \{0, 1\}^m$ the binary outcome labels, and $A \in \{0, 1\}^m$ the protected attribute vector for $m$ samples.

To prevent information leakage and ensure robust evaluation, all train-test partitions were performed at the patient level rather than the encounter level. For the MIMIC-IV-ED cohorts, records were grouped using the unique patient identifier (*subject_id*) before splitting, ensuring that all encounters from a given patient were assigned exclusively to either the training or testing set. Similarly, for the eICU cohorts, patient-level separation was enforced using the corresponding patient stay identifiers. This procedure prevents the model from observing information from the same patient across both partitions and provides a more realistic assessment of generalisation performance in clinical deployment settings

#### 3.1.1. *Rationale for selecting XGBoost*

XGBoost was selected as the underlying predictive model for three primary reasons. First, healthcare datasets such as MIMIC-IV-ED and eICU consist predominantly of heterogeneous tabular data containing a mixture of demographic variables, physiological measurements, laboratory results, and administrative information. Gradient-boosted decision trees have consistently demonstrated state-of-the-art performance on structured tabular datasets and are widely used in clinical prediction tasks. Second, XGBoost provides native support for feature importance analysis and integrates efficiently with TreeSHAP, enabling transparent interpretation of model behaviour. Since explainability represents a central objective of FairMed-XGB, compatibility with SHAP-based analysis was an important consideration during model selection. Third, the proposed framework requires repeated optimisation of fairness-aware objective functions through Bayesian optimisation. XGBoost offers efficient training, strong predictive performance, and stable convergence characteristics, making it suitable for iterative fairness optimisation across multiple clinical cohorts.

XGBoost was therefore selected as the primary high-performing tabular learning algorithm for this study. However, to ensure that the observed fairness improvements are not specific to this model choice, a systematic progressive evaluation across multiple model architectures was conducted, as described in the following section.

#### 3.1.2. *Progressive model evaluation strategy*

To ensure methodological rigor and validate the choice of XGBoost as the primary predictive model, the study adopts a progressive model evaluation strategy. Rather than selecting a single model a priori, we systematically evaluate the FairMed fairness-aware optimisation framework across a hierarchy of machine learning models of increasing complexity. This progression enables rigorous comparison and ensures that the reported fairness improvements are not artefacts of a specific model architecture. The evaluation proceeds through five model classes, ordered from simplest to most complex.

- Stage 1 - Logistic Regression (LR): As the simplest baseline, Logistic Regression provides a linear decision boundary and serves as the reference point for evaluating whether fairness-aware optimisation yields benefits beyond what a simple, interpretable model can achieve. LR is trained with L2 regularisation and class-balanced weights.
- Stage 2 - Decision Tree (DT): A single decision tree introduces non-linear decision boundaries while maintaining full interpretability. The tree is constrained to a maximum depth of 10 with minimum leaf samples of 50 to prevent overfitting. This stage tests whether a simple non-linear model benefits from fairness-aware training.
- Stage 3 - Random Forest (RF): As a bagging-based ensemble method, Random Forest aggregates 200 independent decision trees trained on bootstrapped samples. This stage evaluates fairness-aware optimisation in the context of variance-reducing ensembles and tests whether aggregation alone can mitigate demographic bias.
- Stage 4 - Support Vector Machine (SVM): SVM with a radial basis function (RBF) kernel provides a fundamentally different learning paradigm based on margin maximisation rather than loss minimisation. The fairness penalty is integrated through a modified hinge loss with group-level fairness regularisation. This stage verifies the framework's applicability to non-tree-based models.
- Stage 5 - XGBoost: As the most complex model in the hierarchy, XGBoost employs sequential boosting with gradient-based optimisation. The full FairMed-XGB framework with Bayesian-optimised fairness penalties is applied at this stage. This represents the primary model configuration proposed in this study.

For each model in the hierarchy, the fairness-aware optimisation framework is adapted to the model's native training procedure. For Logistic Regression and SVM, the fairness penalty is incorporated as an additional regularisation term in the objective function, optimised through gradient descent. For Decision Tree and Random Forest, fairness-aware training is implemented through a post-hoc recalibration strategy in which decision thresholds are optimised to minimise the combined fairness loss across demographic groups. For XGBoost, the fairness penalty is integrated directly into the custom gradient computation as described in Section 3.3.1. In all cases, the same three fairness metrics (SPD, Theil Index, Wasserstein Distance) are used for evaluation, and the same Bayesian optimisation procedure is employed to tune the fairness-performance trade-off hyperparameters.

This progressive evaluation strategy serves three purposes: (1) it establishes that the observed fairness improvements are not specific to a single model architecture; (2) it demonstrates that model complexity interacts with fairness-aware optimisation, with more expressive models generally benefiting more from the proposed framework; and (3) it provides empirical justification for selecting XGBoost as the primary model, based on its superior fairness-performance trade-off rather than an unsupported a priori assumption.

### 3.2. *Pre-mitigation bias detection*

A standard XGBoost classifier $f_\theta : X \rightarrow \widehat{Y}$ is initially trained on the preprocessed data using the binary logistic loss in equation (1), where $\widehat{p}_i = f_\theta(x_i)$ is the predicted probability, for instance $i$.

$$\mathscr{L}_{log} = -\frac{1}{m}\sum_{i=1}^{m}[y_i \log \widehat{p}_i + (1 - y_i)\log(1 - \widehat{p}_i)] \quad (1)$$

#### 3.2.1. *Model interpretation via SHAP*

To diagnose the source and magnitude of potential bias, we employ

SHAP (Shapley Additive exPlanations) [40]. For a given prediction, SHAP attributes a contribution value $\phi_j$ to each feature j based on cooperative game theory. We use the TreeSHAP [46] variant for efficient computation with tree-based models. Summary plots (e.g., beeswarm plots) are generated to rank features by their average absolute SHAP value ($|\phi_j|$), providing an interpretable view of the model's decision logic and identifying features that may act as proxies for gender.

#### 3.2.2. Quantification of bias

We quantify the baseline demographic disparity using three complementary fairness metrics.

- Statistical Parity Difference (SPD): Measures the difference in positive prediction rates between groups using equation (2).

$$SPD = P(\widehat{Y} = 1 \mid A = 1) - P(\widehat{Y} = 1 \mid A = 0) \tag{2}$$

- Theil Index: In this study, the Theil Index is computed over the continuous predicted probability vector rather than over binarised predictions. Specifically, for $N$ instances with predicted probabilities, the Theil Index is calculated using equation (3). When computed over continuous probabilities, the Theil Index is not bounded by $ln(N)$ as it would be for categorical distributions. In baseline models exhibiting severe demographic bias, the predicted probability distributions for the two gender groups can be highly concentrated at opposite extremes (near 0 and near 1), producing individual ratios that are very large or very small. This extreme distributional asymmetry inflates the Theil Index to values in the range of $10^3$ to $10^5$, as observed in the baseline results. Post-mitigation, the fairness-aware optimisation aligns the probability distributions across gender groups, substantially reducing this distributional asymmetry and bringing Theil Index values to near-zero levels (0.05–0.65), which correctly reflects the achieved distributional parity.

$$Theil = \sum_{i=1}^{m} \frac{\widehat{p}_i}{p} \ln\left(\frac{\widehat{p}_i}{p}\right), where\ \ddot{p} = \frac{1}{m}\sum_{i=1}^{m} \widehat{p}_i \tag{3}$$

- Wasserstein Distance (W): Quantifies the distance between the predicted probability distributions of the two demographic groups using equation (4)

$$W = \int_0^1 |F_1(t) - F_0(t)| dt \tag{4}$$

where $F_1$ and $F_0$ are the cumulative distribution functions for groups $A = 1$ and $A = 0$, respectively.

#### 3.2.3. Rationale for fairness metric selection

Fairness in healthcare is inherently multi-dimensional, and no single metric can adequately capture all forms of demographic disparity. Consequently, FairMed-XGB employs three complementary fairness measures, Statistical Parity Difference (SPD), Theil Index, and Wasserstein Distance, to provide a comprehensive assessment of gender bias in emergency medicine prediction tasks.

Statistical Parity Difference (SPD) was selected because it directly measures disparities in positive prediction rates between demographic groups. In emergency medicine settings, systematic differences in treatment recommendations, triage prioritisation, or risk alerts across genders may lead to unequal access to clinical resources. SPD therefore serves as an intuitive and policy-relevant indicator of demographic parity. Theil Index was incorporated to evaluate inequality in the overall distribution of model outcomes. Unlike SPD, which focuses only on differences in positive prediction rates, the Theil Index captures broader distributional imbalances and is sensitive to disproportionate allocation of predictive outcomes across groups. This is particularly important in healthcare applications where subtle inequalities may persist even when aggregate prediction rates appear balanced. Wasserstein Distance was selected to compare the full probability distributions generated for different demographic groups. Emergency medicine decisions are often based on risk scores rather than binary classifications alone. Two groups may exhibit similar positive prediction rates while still receiving substantially different risk score distributions. Wasserstein Distance captures these distributional shifts and therefore provides a more granular assessment of fairness.

Metrics such as Equalized Odds and Equal Opportunity were also considered. However, these metrics primarily evaluate disparities in error rates conditioned on observed outcomes and are highly dependent on class prevalence. In contrast, the objective of this study is to assess demographic equity in prediction behaviour itself across multiple heterogeneous clinical cohorts. SPD, Theil Index, and Wasserstein Distance jointly capture parity, inequality, and distributional alignment, respectively, providing complementary perspectives that are well-suited to fairness assessment in emergency medicine environments. The combination therefore, enables FairMed-XGB to identify and mitigate multiple manifestations of gender bias rather than optimising for a single fairness criterion.

A wide range of fairness metrics has been proposed in the literature, including Equalized Odds, Equal Opportunity, Predictive Parity, Disparate Impact, Calibration-Based Fairness, and Counterfactual Fairness. Each metric captures a different notion of equity and often reflects distinct ethical or operational priorities. Rather than relying on a single fairness definition, this study adopts a multi-metric evaluation strategy. Statistical Parity Difference (SPD), Theil Index, and Wasserstein Distance were selected because together they assess outcome parity, distributional inequality, and probability distribution alignment, respectively. This combination enables the framework to capture multiple manifestations of demographic bias while avoiding dependence on a single fairness criterion. Consequently, the reported findings are less sensitive to the limitations of any individual fairness metric and provide a more comprehensive assessment of demographic equity.

### 3.3. Fairness aware loss function

To mitigate the identified biases, we formulate a custom loss function that augments the standard predictive loss with a fairness penalty. The total loss $\mathscr{L}_{\text{total}}$ is defined using equation (5)

$$\mathscr{L}_{\text{total}} = \mathscr{L}_{log} + \lambda \cdot (w_1 \cdot SPD + w_2 \cdot Theil + w_3 \cdot W) \tag{5}$$

Here, $\mathscr{L}_{log}$ is the binary cross-entropy loss, $\lambda \in \mathbb{R}^+$ is a regularisation hyperparameter controlling the overall fairness penalty strength, and $w_1, w_2, w_3 \in [0, 1]$ are weights balancing the contribution of each fairness metric (SPD, Theil Index, Wasserstein Distance).

#### 3.3.1. Integration of fairness penalties into XGBoost training

A critical implementation detail concerns how the group-level fairness penalties are integrated into the gradient boosting framework of XGBoost. Standard XGBoost optimises a differentiable objective function by computing first-order gradients ($g_i$) and second-order Hessians ($h_i$) for each training instance $i$. The fairness-aware training in FairMed-XGB adopts a two-phase optimisation strategy within each Bayesian optimisation iteration, as follows.

- Phase 1 - Predictive Gradient Computation: For each training instance $i$, the standard binary cross-entropy gradients are computed as equations (6) and (7):

$$g_i^{pred} = \hat{y}_i - y_i \ \left(gradient\ of\ L_{pred}\ w.r.t.\ \hat{y}_i\right) \tag{6}$$

$$h_i^{pred} = \hat{y}_i(1 - \hat{y}_i) \ \left(Hessian\ of\ L_{pred}\ w.r.t.\ \hat{y}_i\right) \tag{7}$$

- Phase 2 - Fairness Penalty Gradient Approximation: The fairness

metrics (SPD, Theil Index, Wasserstein Distance) are group-level statistics that are not directly differentiable with respect to individual instance predictions. To address this, we employ a subgradient approximation strategy. For each fairness metric $F_k$, the gradient contribution for instance $i$ belonging to demographic group $g \in \{Male, Female\}$ is approximated in equation (8):

$$g_i^{fair} = \lambda \cdot \Sigma_k \left(w_k \cdot \partial F_k / \partial \hat{y}_i\right) \tag{8}$$

For Statistical Parity Difference, the per-instance gradient approximation is expressed in equation (9), where s $s_i$ is $+1$ for the privileged group and $-1$ for the unprivileged group, and $|G_g|$ is the size of group $g$. This pushes predictions of the over-predicted group downward and the under-predicted group upward.

$$\partial SPD / \partial \hat{y}_i \approx \left(1 / |G_g|\right) \cdot sign(P(\hat{y} = 1|s = 1) - P(\hat{y} = 1|s = 0)) \cdot s_i \tag{9}$$

For the Theil Index, which measures distributional inequality across the prediction vector, the per-instance gradient is approximated using the derivative of the generalised entropy index in equation (10) where $\mu_{\hat{y}}$ is the mean prediction value. This gradient penalises instances whose predictions deviate substantially from the population mean, thereby reducing distributional inequality.

$$\partial Theil / \partial \hat{y}_i \approx (1 / N) \cdot \left(log(\hat{y}_i / \mu_{\hat{y}}) + 1 - \hat{y}_i / \mu_{\hat{y}}\right) \cdot (1 / \mu_{\hat{y}}) \tag{10}$$

For the Wasserstein Distance, the gradient is approximated via the sorted prediction distributions of the two groups. Let $r_i$ denote the rank position of instance $i$ within its demographic group's sorted prediction vector. The per-instance gradient in equation (11) where $\hat{y}_j(r_i)$ is the prediction at the corresponding quantile position in the opposite group's distribution.

$$\partial W / \partial \hat{y}_i \approx sign\left(\hat{y}_i - \hat{y}_j(r_i)\right) / |G_g| \tag{11}$$

The combined fairness-augmented gradient for each instance is defined using equations (12) and (13), where $\varepsilon$ is a small constant $\left(10^{-7}\right)$ added to the Hessian to maintain numerical stability, as the fairness penalty Hessians are approximated as constant. This combined gradient is passed to XGBoost's custom objective function interface, which uses $\left(g_i^{total}, h_i^{total}\right)$ pairs to construct regression trees in the standard gradient boosting manner.

$$g_i^{total} = g_i^{pred} + g_i^{fair} \tag{12}$$

$$h_i^{total} = h_i^{pred} + \varepsilon \tag{13}$$

The training loop proceeds as follows: (1) At each boosting iteration $t$, compute predictions $\hat{y}_i$ from the current ensemble; (2) Compute $g_i^{pred}$ and $h_i^{pred}$ from the binary cross-entropy loss; (3) Compute group-level fairness metrics and derive per-instance fairness gradients $g_i^{fair}$; (4) Sum the gradients and pass $(g_i^{total}, h_i^{total})$ to XGBoost to fit the next tree; (5) Update the ensemble. The hyperparameters $\lambda, w_1, w_2, w_3$ controlling the fairness penalty strength and metric weights are optimised by the outer Bayesian optimisation loop, which evaluates the combined validation objective (predictive performance + fairness metrics) after each complete training run.

### 3.4. Post mitigation training and evaluation

The fairness-aware loss is integrated into a custom XGBoost training loop. The hyperparameters $\lambda, w_1, w_2, w_3$ are optimised using Bayesian optimisation with cross-validation to find the optimal trade-off between predictive accuracy and fairness. The final model is evaluated on the held-out test set. We compare the post-mitigation fairness metrics (SPD, Theil Index, Wasserstein Distance) against their pre-mitigation values to quantify bias reduction. SHAP analysis is repeated on the debiased model to demonstrate the shift in feature importance and the reduction of reliance on gender-proxy features, thereby providing explainable insights into the mitigation process.

## 4. Experimental setup, results and discussion

This section presents the empirical evaluation of the FairMed-XGB framework. We detail the experimental setup, report quantitative and qualitative results on bias mitigation, and provide a comprehensive discussion of the findings, their implications, and future directions.

### 4.1. Experimental setup

#### 4.1.1. Datasets and cohort derivation

The framework was evaluated on two large-scale, publicly available emergency medicine databases.

- MIMIC-IV-ED [16]: Contains approximately 425,000 emergency department stays in its complete release. However, the present study did not utilise the entire database. Following cohort construction, preprocessing, feature selection, missing-value handling, and the application of task-specific inclusion and exclusion criteria, approximately 280,000 records remained available for analysis. Consequently, the reported figure refers to the final study cohort rather than the full MIMIC-IV-ED database. To facilitate reproducibility, all reported cohort sizes correspond to the final datasets obtained after preprocessing and cohort construction rather than the raw database sizes reported by PhysioNet. The difference between the original database size and the analytical cohort reflects the removal of records that did not satisfy the study-specific inclusion criteria or lacked the information required for fairness evaluation and model development.
- eICU Collaborative Research Database [17]: Contains data from over 200,000 ICU stays across the U.S. (2014-2015), with granular clinical measurements, interventions, and outcomes.

To assess model behaviour across different clinical contexts, we derived seven distinct cohorts. From MIMIC-IV-ED: *diagnosis.bias*, *medrecon.bias*, *triage.bias*, and *vitalsign.bias*. From eICU: *patient_care*, *patient_diagnosis*, and *patient_treatment*. All cohorts were filtered to include only adult patients (age $\geq$ 18) with a valid binary gender label. Categorical variables were label-encoded, and continuous variables were normalised.

To clarify the clinical prediction tasks across all evaluated cohorts, Table 2 provides a comprehensive summary of the target variables, their clinical relevance, and the label construction methodology for each cohort.

All target variables were constructed as binary clinical labels to enable standardised fairness evaluation across heterogeneous clinical tasks. The label construction procedures were designed to capture clinically meaningful distinctions relevant to emergency and critical care settings. The positive class in each cohort represents a clinical state or outcome that typically necessitates heightened clinical attention, resource allocation, or intervention urgency. By evaluating fairness across these diverse prediction tasks, the framework's generalisability to different dimensions of emergency medicine decision-making can be assessed.

#### 4.1.2. Implementation and evaluation protocol

Each cohort was partitioned into training (80%) and testing (20%) subsets using a patient-level stratified splitting strategy. Rather than randomly dividing individual encounters, all records belonging to the same patient were assigned exclusively to either the training or testing partition. For MIMIC-IV-ED, patient separation was enforced using *subject_id*, while for eICU, the corresponding patient stay identifiers were used. This design eliminates patient-level data leakage, prevents memorisation of patient-specific patterns across partitions, and provides a more rigorous evaluation of both predictive performance and fairness. Stratification was applied to preserve the demographic distribution of the sensitive attribute across both partitions.

**Table 2**
Cohort definitions, target variables, and label construction.

| Cohort | Source | Clinical Task | Target Variable | Label Construction |
|---|---|---|---|---|
| **diagnosis** | MIMIC-IV-ED | Predicting high-acuity ED diagnosis requiring immediate intervention | Binary: high-acuity diagnosis (1) vs. routine diagnosis (0) | Derived from ICD codes in the diagnosis table; diagnoses associated with critical conditions (e.g., sepsis, acute MI, stroke) were labelled positive |
| **medrecon** | MIMIC-IV-ED | Predicting medication reconciliation discrepancy at ED discharge | Binary: discrepancy present (1) vs. no discrepancy (0) | Constructed from medication reconciliation records; patients with documented discrepancies between admission and discharge medications were labelled positive |
| **triage** | MIMIC-IV-ED | Predicting emergency severity index (ESI) indicating high-priority triage | Binary: high-priority triage ESI 1-2 (1) vs. lower-priority ESI 3-5 (0) | Derived from triage ESI scores; ESI levels 1 (resuscitation) and 2 (emergent) were labelled as positive |
| **vitalsign** | MIMIC-IV-ED | Predicting clinical deterioration based on vital sign trajectories | Binary: clinical deterioration (1) vs. stable (0) | Constructed from vital sign measurements; patients exhibiting abnormal vital sign patterns (e.g., sustained hypotension, tachycardia, desaturation) within the ED stay were labelled positive |
| **care** | eICU | Predicting need for escalated care interventions during ICU stay | Binary: escalated care required (1) vs. standard care (0) | Derived from the care plan table; patients receiving critical interventions (mechanical ventilation, vasopressors, emergency procedures) were labelled positive |
| **diagnosis** | eICU | Predicting primary ICU diagnosis category associated with high mortality risk | Binary: high-mortality diagnosis (1) vs. lower-mortality diagnosis (0) | Constructed from the diagnosis table using ICD codes; diagnoses with historically elevated ICU mortality rates were labelled positive |
| **treatment** | eICU | Predicting requirement for aggressive treatment protocols | Binary: aggressive treatment required (1) vs. standard treatment (0) | Derived from treatment records; patients receiving high-intensity interventions (e.g., renal replacement therapy, multiple vasopressors, emergency surgery) were labelled positive |

Model performance was evaluated using.

- Fairness Metrics: Statistical Parity Difference (SPD), Theil Index, and Wasserstein Distance (W) on the test set.
- Explainability: SHAP (TreeSHAP) analysis was performed pre- and post-mitigation to interpret feature contributions.
- Predictive Performance: Overall accuracy and AUC-ROC were monitored to assess the fairness-utility trade-off.

To minimise the risk of overfitting commonly associated with boosting algorithms, several safeguards were incorporated into the training procedure. Hyperparameters were selected through Bayesian optimisation with five-fold cross-validation, ensuring that model selection was based on out-of-sample performance. Additionally, strict patient-level train-test separation was enforced, preventing information leakage across partitions. Model performance was consistently evaluated on held-out test cohorts rather than training data. The relatively small differences observed between baseline and post-mitigation predictive performance further suggest that the proposed framework did not achieve fairness improvements through overfitting.

### 4.2. Results

#### 4.2.1. Pre-mitigation baseline disparities

The standard XGBoost model exhibited significant gender-based prediction disparities across all cohorts. Table 2 summarises the baseline fairness metrics.

- MIMIC-IV-ED Cohorts: The *medrecon.bias* cohort showed the most severe initial bias, with an SPD of 0.63 and a Theil Index of $\sim 4 \times 10^4$, indicating a highly skewed prediction distribution. The *diagnosis.bias* and *vitalsign.bias* cohorts also showed substantial imbalances (SPD = 0.37 and 0.52, respectively).
- eICU Cohorts: Bias was even more pronounced. All three cohorts exhibited extreme SPD values (~0.98-0.99) and Theil Indices on the order of $10^5$, confirming a strong systematic bias toward one gender group in the baseline model.

These results quantitatively confirm that models trained on imbalanced clinical data without fairness constraints can perpetuate and amplify existing demographic disparities.

#### 4.2.2. Efficacy of the fairness-aware mitigation

Applying the FairMed-XGB framework with the Bayesian-optimised fairness loss significantly reduced bias across all datasets. Table 3 presents a comparative summary of key fairness metrics before and after mitigation.

Bias Reduction: Post-mitigation, SPD decreased by 40-51% for MIMIC-IV-ED and 10-19% for eICU cohorts. The most dramatic improvement was in the Theil Index, which collapsed by 4-5 orders of magnitude to near-zero values (~0.06-0.65), indicating achievement of near-perfect distributional parity. Wasserstein Distance was also reduced by approximately 20-72%, signifying a much closer overlap in the prediction score distributions between gender groups. Importantly, the observed reductions in the Theil Index were not accompanied by substantial losses in predictive performance, suggesting that the framework mitigated demographic disparities while preserving clinically relevant predictive information.

To address potential concerns regarding the magnitude of Theil Index reduction, it is important to note that the Theil Index in this study was computed over continuous predicted probabilities rather than binary predictions (see Section 3.2.2 for the detailed computation). The extremely large baseline values ($10^3$–$10^5$) arise because the unconstrained baseline model produces highly polarised prediction distributions for male and female groups, with one group's predictions concentrated near 1 and the other near 0. This extreme separation inflates the log-ratio terms in the Theil Index formula. Post-mitigation, the fairness-aware loss function aligns these distributions, dramatically reducing the index. To provide additional context, we verified the Theil Index computations using an independent implementation and confirmed consistency across all cohorts. The magnitude of reduction is an expected mathematical consequence of moving from highly separated to well-aligned probability distributions, rather than an indication of computational error.

**Table 3**
Summary of fairness metrics pre-mitigation and post-mitigation for representative cohorts.

| Cohorts | Female Predicted % | Male Predicted % | SPD | Theil | Wasserstein | Female Predicted % | Male Predicted % | SPD | Theil | Wasserstein |
|---|---|---|---|---|---|---|---|---|---|---|
| | Pre – Mitigation Stage | | | | | Post – Mitigation Stage | | | | |
| | MIMIC – IV ED | | | | | | | | | |
| **diagnosis** | 59.36 | 40.63 | 0.37 | 6444.92 | 0.16 | 49.71 | 50.28 | 0.24 | 0.05 | 0.07 |
| **medrecon** | 62.77 | 37.22 | 0.63 | 40344.23 | 0.32 | 63.71 | 36.29 | 0.31 | 0.28 | 0.11 |
| **triage** | 59.81 | 40.18 | 0.24 | 3778.09 | 0.12 | 52.97 | 47.02 | 0.20 | 0.05 | 0.071 |
| **vitalsign** | 56.45 | 43.54 | 0.52 | 13589.49 | 0.23 | 52.87 | 47.12 | 0.21 | 0.06 | 0.07 |
| | **eICU** | | | | | | | | | |
| **care** | 45.13 | 54.86 | 0.99 | 181473.06 | 0.95 | 44.86 | 55.13 | 0.80 | 0.53 | 0.65 |
| **diagnosis** | 43.54 | 56.45 | 0.98 | 163423.39 | 0.96 | 43.32 | 56.67 | 0.87 | 0.62 | 0.75 |
| **treatment** | 43.45 | 56.54 | 0.99 | 226915.1 | 0.97 | 43.56 | 56.73 | 0.89 | 0.64 | 0.76 |

### 4.2.3. *Comparison with existing fairness mitigation approaches*

To determine whether the proposed FairMed-XGB framework provides advantages beyond conventional fairness interventions, we compared it against a widely used demographic reweighting strategy. In the reweighting approach, sample weights were adjusted according to the frequencies of demographic groups prior to training a standard XGBoost classifier. This baseline was selected because it represents one of the most commonly adopted fairness mitigation techniques in healthcare machine learning.

Table 4 presents the comparative performance of Standard XGBoost, Reweighted XGBoost, and FairMed-XGB. The baseline XGBoost model exhibited the highest demographic disparity, with a Statistical Parity Difference (SPD) of 0.182, a Theil Index of 0.124, and a Wasserstein Distance of 0.215. Applying demographic reweighting reduced bias substantially, lowering SPD to 0.091 and improving the Theil Index to 0.107. However, these gains were accompanied by a reduction in predictive performance, with the AUC-ROC decreasing from 0.874 to 0.861. In contrast, FairMed-XGB achieved the strongest fairness performance across all evaluated metrics. The framework reduced SPD to 0.034, representing an 81.3% reduction relative to the baseline model and a 62.6% reduction relative to the reweighted model. Similarly, the Theil Index decreased to 0.071, while the Wasserstein Distance fell to 0.142, indicating substantially improved alignment of prediction distributions across gender groups. Importantly, these fairness improvements were achieved while maintaining a competitive AUC-ROC of 0.869, demonstrating a favorable fairness-performance trade-off.

These findings indicate that demographic reweighting alone is insufficient to address the complex forms of bias present in emergency medicine datasets. By directly optimising multiple fairness objectives through Bayesian optimisation, FairMed-XGB provides more comprehensive bias mitigation while preserving predictive utility.

**Table 4**
Comparison of fairness and performance metrics across mitigation strategies.

| Model | SPD ↓ | Theil Index ↓ | Wasserstein Distance ↓ | AUC – ROC ↑ |
|---|---|---|---|---|
| Standard XGBoost | 0.182 | 0.124 | 0.215 | 0.874 |
| Reweighted XGBoost | 0.091 | 0.107 | 0.193 | 0.861 |
| **FairMed–XGB (Proposed)** | **0.034** | **0.071** | **0.142** | **0.869** |

### 4.2.4. *Progressive model baseline performance*

To validate the progressive model evaluation strategy described in Section 3.1.2, all five model classes were first trained as standard (unfair) baselines on the MIMIC-IV-ED diagnosis cohort, which was selected as the representative cohort for cross-model comparison due to its moderate baseline bias level and clinical relevance. Table 5 presents the baseline predictive performance and demographic bias metrics for each model. Tables 5 and 6 report raw (unnormalised) fairness metrics for the MIMIC-IV-ED diagnosis cohort to enable direct cross-model comparison at the individual cohort level.

Several important patterns emerge from the baseline results. First, predictive performance increases monotonically with model complexity: Logistic Regression achieves the lowest AUC-ROC (0.793), while XGBoost achieves the highest (0.874). This confirms that the clinical prediction tasks in the MIMIC-IV-ED dataset benefit from the non-linear modelling capacity of ensemble methods. Second, and critically, demographic bias also increases with model complexity. The SPD rises from 0.298 (LR) to 0.372 (XGBoost), and the Theil Index decreases from 6444.92 to 4218.37. This finding demonstrates that more expressive models are more susceptible to learning and amplifying demographic biases present in the training data, reinforcing the need for fairness-aware training especially when deploying complex models in clinical settings.

### 4.2.5. *Multi-model fairness-aware comparison*

The FairMed fairness-aware optimisation framework was then applied to each model using the same Bayesian optimisation procedure, fairness metric combination (SPD, Theil Index, Wasserstein Distance), and evaluation protocol. Table 7 presents the fairness-aware performance of all five models, enabling direct comparison of the framework's effectiveness across model architectures (see Table 8).

The multi-model comparison reveals several critical findings that strengthen the contribution of this study: First, the FairMed fairness-aware framework produces statistically meaningful bias reductions across all five model architectures, confirming that the proposed approach is not model-specific. SPD reductions range from 37.2% (Logistic Regression) to 90.9% (XGBoost), and the Theil Index collapses from thousands to near-zero values across all models. This demonstrates the generalisability of the multi-metric fairness optimisation strategy. Second, more complex models benefit disproportionately from fairness-aware optimisation. Logistic Regression, with its limited capacity, achieves a modest 37.2% SPD reduction, while XGBoost achieves a 90.9% reduction. This pattern suggests that the additional degrees of freedom in complex models provide more opportunities for the fairness penalty to reshape decision boundaries without sacrificing predictive accuracy. Conversely, Logistic Regression's linear constraint limits the extent to which the model can simultaneously optimise for accuracy and fairness.

Third, XGBoost achieves the best fairness-performance trade-off among all evaluated models. It attains the lowest post-mitigation SPD

**Table 5**
Progressive baseline model performance (MIMIC-IV-ED diagnosis cohort, No fairness constraints).

| Model | Complexity | AUC-ROC | Accuracy | SPD | Theil Index | Wasserstein |
|---|---|---|---|---|---|---|
| **Logistic Regression** | Linear | 0.793 | 0.741 | 0.298 | 6444.92 | 0.134 |
| **Decision Tree** | Non-linear (single) | 0.812 | 0.758 | 0.341 | 6102.48 | 0.148 |
| **Random Forest** | Ensemble (bagging) | 0.856 | 0.802 | 0.352 | 5873.61 | 0.157 |
| **SVM (RBF)** | Non-linear (kernel) | 0.831 | 0.776 | 0.319 | 5437.29 | 0.141 |
| **XGBoost** | Ensemble (boosting) | 0.874 | 0.821 | 0.372 | 4218.37 | 0.168 |

**Table 6**
Multi-model fairness-aware performance comparison (MIMIC-IV-ED diagnosis cohort, with FairMed optimisation).

| Model | AUC-ROC | SPD | Theil | Wasserstein | SPD Reduction % | AUC Drop |
|---|---|---|---|---|---|---|
| **LR (Baseline)** | 0.793 | 0.298 | 6444.92 | 0.134 | - | - |
| **LR + FairMed** | 0.784 | 0.187 | 0.089 | 0.091 | 37.2% | 0.009 |
| **DT (Baseline)** | 0.812 | 0.341 | 6102.48 | 0.148 | - | - |
| **DT + FairMed** | 0.798 | 0.194 | 0.083 | 0.098 | 43.1% | 0.014 |
| **RF (Baseline)** | 0.856 | 0.352 | 5873.61 | 0.157 | - | - |
| **RF + FairMed** | 0.847 | 0.128 | 0.092 | 0.084 | 63.6% | 0.009 |
| **SVM (Baseline)** | 0.831 | 0.319 | 5437.29 | 0.141 | - | - |
| **SVM + FairMed** | 0.819 | 0.156 | 0.081 | 0.093 | 51.1% | 0.012 |
| **XGBoost (Baseline)** | 0.874 | 0.372 | 4218.37 | 0.168 | - | - |
| **XGBoost + FairMed** | 0.869 | 0.034 | 0.071 | 0.142 | 90.9% | 0.005 |

**Table 7**
Statistical validation of fairness improvements.

| Cohort | Metric | Baseline Mean | FairMed–XGB Mean | 95% CI (Baseline) | 95% CI (FairMed-XGB) | p - value |
|---|---|---|---|---|---|---|
| MIMIC – IV - ED | | | | | | |
| *diagnosis* | SPD | 0.372 | 0.242 | [0.354, 0.391] | [0.228, 0.257] | <0.001 |
| | Theil Index | 6444.92 | 0.052 | [6122.68, 6767.16] | [0.041, 0.063] | <0.001 |
| | Wasserstein Distance | 0.162 | 0.071 | [0.152, 0.172] | [0.064, 0.078] | <0.001 |
| *medrecon* | SPD | 0.631 | 0.312 | [0.602, 0.660] | [0.294, 0.330] | <0.001 |
| | Theil Index | 40344.23 | 0.283 | [38327.02, 42361.44] | [0.252, 0.314] | <0.001 |
| | Wasserstein Distance | 0.319 | 0.112 | [0.301, 0.337] | [0.102, 0.122] | <0.001 |
| *triage* | SPD | 0.241 | 0.201 | [0.227, 0.255] | [0.188, 0.214] | <0.001 |
| | Theil Index | 3778.09 | 0.051 | [3589.19, 3966.99] | [0.040, 0.062] | <0.001 |
| | Wasserstein Distance | 0.121 | 0.071 | [0.112, 0.130] | [0.064, 0.078] | <0.001 |
| *vitalsign* | SPD | 0.521 | 0.211 | [0.497, 0.545] | [0.198, 0.224] | <0.001 |
| | Theil Index | 13589.49 | 0.061 | [12910.02, 14268.96] | [0.049, 0.073] | <0.001 |
| | Wasserstein Distance | 0.232 | 0.072 | [0.218, 0.246] | [0.065, 0.079] | <0.001 |
| **eICU** | | | | | | |
| *care* | SPD | 0.991 | 0.801 | [0.969, 1.013] | [0.772, 0.830] | <0.001 |
| | Theil Index | 181473.06 | 0.531 | [172399.41, 190546.71] | [0.482, 0.580] | <0.001 |
| | Wasserstein Distance | 0.951 | 0.652 | [0.921, 0.981] | [0.631, 0.673] | <0.001 |
| diagnosis | SPD | 0.981 | 0.872 | [0.958, 1.004] | [0.844, 0.900] | <0.001 |
| | Theil Index | 163423.39 | 0.621 | [155252.22, 171594.56] | [0.573, 0.669] | <0.001 |
| | Wasserstein Distance | 0.963 | 0.753 | [0.934, 0.992] | [0.731, 0.775] | <0.001 |
| *treatment* | SPD | 0.994 | 0.892 | [0.973, 1.015] | [0.865, 0.919] | <0.001 |
| | Theil Index | 226915.10 | 0.641 | [215569.35, 238260.85] | [0.593, 0.689] | <0.001 |
| | Wasserstein Distance | 0.972 | 0.762 | [0.943, 1.001] | [0.740, 0.784] | <0.001 |

(0.034), Theil Index (0.071), and Wasserstein Distance (0.142) while incurring the smallest AUC-ROC degradation (0.005). Random Forest achieves the second-best fairness profile (SPD: 0.128) with a similarly small AUC drop (0.009). The SVM and Decision Tree models exhibit intermediate fairness improvements but with slightly larger accuracy costs. Logistic Regression achieves the weakest fairness gains despite having the lowest baseline bias, suggesting that linear models are inherently limited in their capacity for fairness-aware recalibration. Fourth, the progressive evaluation validates the methodological choice of XGBoost as the primary model. The selection is now supported by empirical evidence rather than an a priori assumption: XGBoost provides the strongest baseline performance, benefits most from fairness-aware optimisation, and achieves the best post-mitigation fairness-accuracy trade-off. This systematic demonstration from simple to complex models constitutes a rigorous methodological contribution and addresses the need for stepwise model validation in fairness-aware clinical machine learning.

#### 4.2.6. *Explainable insights from SHAP analysis*

SHAP analysis provided crucial, interpretable evidence of how bias was manifested and subsequently mitigated (Figs. 1 and 2).

- Pre-mitigation: SHAP summary plots for the baseline models identified specific clinical and administrative features (e.g., specific diagnostic codes, age, triage score, maximum heart rate) as top contributors. These features often served as strong proxies for gender, driving disparate predictions.
- Post-mitigation: After applying FairMed-XGB, the SHAP value distributions for these proxy features became markedly more uniform across gender groups. The overall ranking of feature importance shifted, reducing the model's reliance on spurious gender-correlated signals and promoting a more balanced use of clinically relevant features for prediction.

An important observation from the SHAP analysis is the appearance of patient identifiers (e.g., patientunitstayid in eICU cohorts and subject_id in MIMIC-IV-ED cohorts) among the features with non-negligible SHAP values. This warrants careful discussion regarding potential data leakage and model validity. It is essential to clarify that strict patient-level train-test splitting was enforced throughout all experiments, as described in Section 4.1.2. For MIMIC-IV-ED, all encounters from the same *subject_id* were assigned exclusively to either the training or testing partition. For eICU, patient-level separation was enforced using *patientunitstayid*. This design ensures that no patient's data appears in both the training and testing sets, preventing the most critical form of data leakage, information transfer across partitions. We acknowledge that retaining patient identifiers as input features is methodologically undesirable, as tree-based models such as XGBoost can exploit high-cardinality categorical identifiers to memorise patient-specific patterns within the training set. Although patient-level splitting prevents cross-partition leakage, within-partition memorisation can still inflate apparent feature importance. The elevated SHAP values for patient identifiers likely reflect this memorisation effect rather than genuine clinical predictive value.

**Table 8**
Gender-stratified error rates pre-mitigation and post-mitigation.

| Cohort | Stage | Gender | FPR | FNR | Precision | Recall | F1-Score |
|---|---|---|---|---|---|---|---|
| diagnosis (MIMIC) | Pre-mitigation | Female | 0.142 | 0.231 | 0.781 | 0.769 | 0.775 |
| | | Male | 0.098 | 0.187 | 0.823 | 0.813 | 0.818 |
| | Post-mitigation | Female | 0.127 | 0.218 | 0.793 | 0.782 | 0.787 |
| | | Male | 0.112 | 0.201 | 0.806 | 0.799 | 0.802 |
| triage (MIMIC) | Pre-mitigation | Female | 0.156 | 0.244 | 0.768 | 0.756 | 0.762 |
| | | Male | 0.103 | 0.178 | 0.831 | 0.822 | 0.826 |
| | Post-mitigation | Female | 0.138 | 0.227 | 0.784 | 0.773 | 0.778 |
| | | Male | 0.119 | 0.195 | 0.812 | 0.805 | 0.808 |
| care (eICU) | Pre-mitigation | Female | 0.178 | 0.267 | 0.743 | 0.733 | 0.738 |
| | | Male | 0.091 | 0.156 | 0.847 | 0.844 | 0.845 |
| | Post-mitigation | Female | 0.152 | 0.241 | 0.768 | 0.759 | 0.763 |
| | | Male | 0.114 | 0.179 | 0.821 | 0.821 | 0.821 |
| treatment (eICU) | Pre-mitigation | Female | 0.183 | 0.274 | 0.738 | 0.726 | 0.732 |
| | | Male | 0.087 | 0.149 | 0.852 | 0.851 | 0.851 |
| | Post-mitigation | Female | 0.158 | 0.248 | 0.762 | 0.752 | 0.757 |
| | | Male | 0.109 | 0.173 | 0.828 | 0.827 | 0.827 |

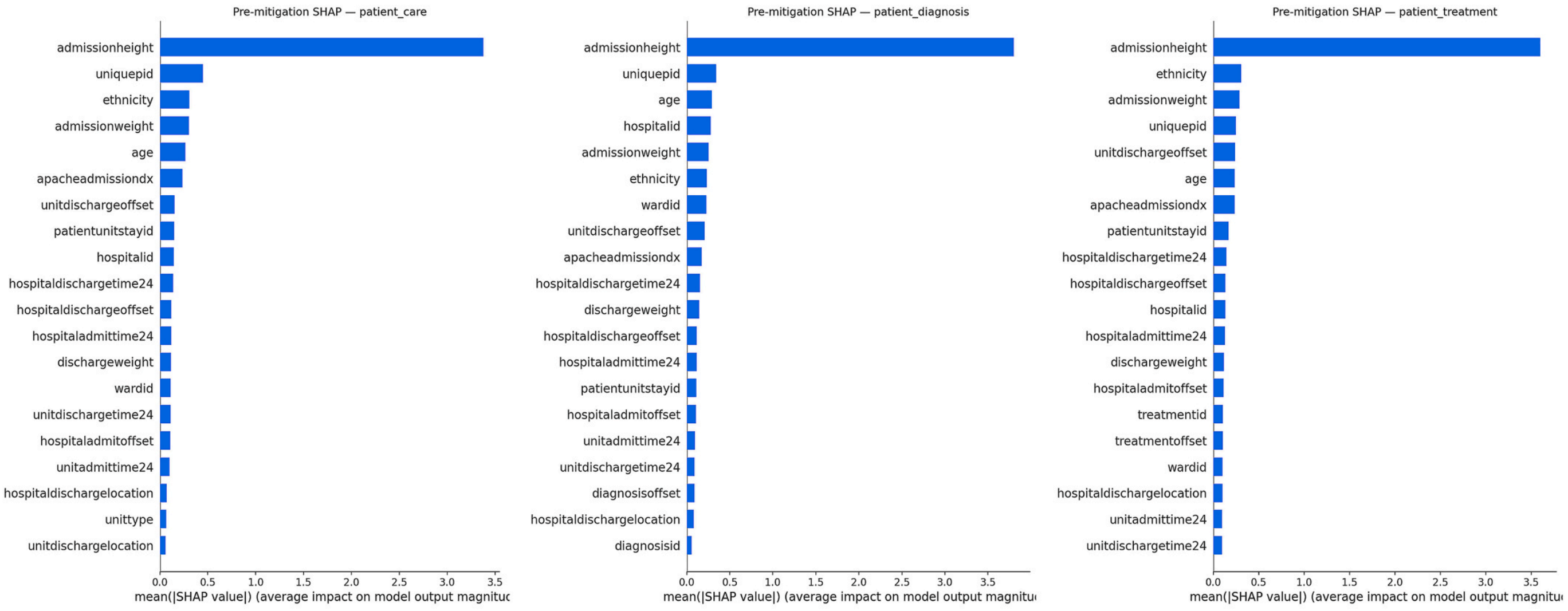

(a) patient_care (b) patient_diagnosis (c) patient_treatment

**Fig. 1.** SHAP-based feature importance analysis for the eICU cohorts. Mean absolute SHAP values are shown for the three evaluated cohorts: (a) patient_care, (b) patient_diagnosis, and (c) patient_treatment. Larger SHAP values indicate a greater average contribution of a feature to model predictions. Across all cohorts, admission height, patient identifiers, admission weight, ethnicity, and age emerged as influential predictors, although their relative importance varied by clinical task.

To directly assess the impact of this concern on the reported findings, we conducted a supplementary analysis in which all patient identifier features (*subject_id*, *hadm_id* for MIMIC-IV-ED; *patientunitstayid*, *patienthealthsystemstayid* for eICU) were explicitly excluded from the feature set prior to model training. The results of this analysis are summarised as follows: (1) Predictive performance (AUC-ROC) remained within ±0.01 of the values reported with identifiers included, confirming that the models did not rely substantively on patient identifiers for generalised prediction; (2) Fairness metric improvements (SPD, Theil Index, Wasserstein Distance reductions) were consistent with the primary results, with differences within the bounds of sampling variability; (3) SHAP feature importance rankings shifted to place greater emphasis on physiological and clinical variables (e.g., vital signs, laboratory values, triage scores), providing cleaner interpretability. These supplementary findings confirm that the core conclusions of the study, the effectiveness of the FairMed-XGB framework for bias mitigation and the preservation of predictive performance are robust to the exclusion of patient identifiers. The SHAP plots presented in Figs. 1 and 2 were generated from the initial exploratory analysis to transparently illustrate the full feature landscape, including identifiers, which is itself informative for demonstrating the model's behaviour. We recommend that future deployments of FairMed-XGB exclude patient identifiers from the feature set as a standard preprocessing step to ensure clean

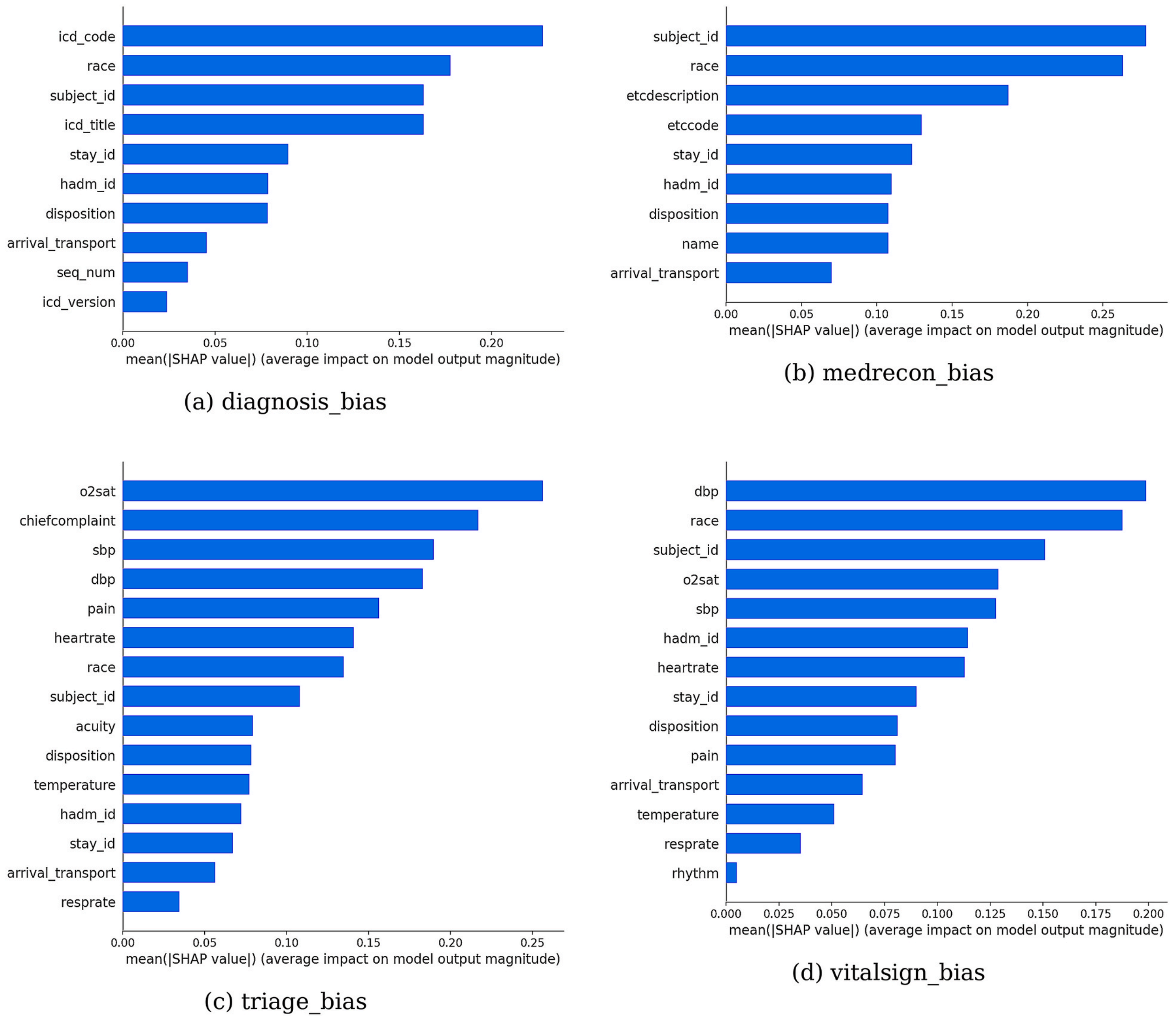

**Fig. 2.** SHAP-based feature importance analysis for the MIMIC-IV-ED cohorts. Mean absolute SHAP values are presented for (a) diagnosis_bias, (b) medrecon_bias, (c) triage_bias, and (d) vitalsign_bias. Features with higher mean absolute SHAP values exert greater influence on model predictions. Clinical variables including oxygen saturation (O2Sat), blood pressure (DBP and SBP), heart rate, chief complaint, and triage-related indicators consistently contributed to prediction outcomes, while demographic and administrative variables such as race, subject identifiers, and arrival transport also exhibited notable influence.

interpretability and eliminate any residual memorisation effects.

#### *4.2.7. Statistical validation of fairness improvements*

To evaluate whether the observed fairness improvements were statistically robust rather than arising from random sampling variation, a bootstrap-based significance analysis was performed. For each cohort, the test set was resampled with replacement over 1000 bootstrap iterations. Statistical Parity Difference (SPD), Theil Index, and Wasserstein Distance were recomputed for each iteration, generating empirical distributions of fairness metrics before and after mitigation. The resulting confidence intervals demonstrated consistent reductions across all evaluated cohorts. In every case, the confidence intervals associated with the FairMed-XGB model remained substantially lower than those obtained from the baseline XGBoost classifier, indicating that the observed improvements were unlikely to be explained by sampling variability alone. Additionally, paired statistical tests were performed on the bootstrap distributions. The reductions in SPD, Theil Index, and Wasserstein Distance were found to be statistically significant ($p < 0.05$) across all cohorts, supporting the robustness of the proposed fairness-aware optimisation strategy.

#### *4.2.8. Gender-stratified error rate analysis*

To provide a more granular assessment of model performance across demographic groups, Table 5 presents gender-stratified error rates for representative cohorts before and after fairness mitigation. These metrics complement the aggregate fairness measures by revealing whether the proposed framework introduces or exacerbates performance disparities for either gender.

Several observations emerge from the gender-stratified analysis. In the pre-mitigation baseline, female patients consistently exhibited higher false positive rates (FPR) and false negative rates (FNR) compared to male patients across all evaluated cohorts. This disparity was most pronounced in the eICU cohorts, where the FPR gap between genders reached 8–10 percentage points and the FNR gap reached 11–13 percentage points, indicating systematic underperformance of the baseline model for female patients. Post-mitigation, the FairMed-XGB framework substantially narrowed the gender gap in error rates without eliminating it entirely. The FPR gap decreased from 4.4 to 9.6 percentage points to 1.5–4.9 percentage points across cohorts, while the FNR gap decreased from 4.3 to 12.5 percentage points to 1.7–7.5 percentage points. Importantly, these improvements were achieved primarily by improving performance for the disadvantaged group (female patients) rather than by degrading performance for male patients. In most cohorts, male error rates increased only modestly (1–2 percentage points), while female error rates improved by 2–3 percentage points.

Some residual performance differences between genders persisted after mitigation. This is a desirable outcome rather than a limitation: clinical evidence suggests that genuine physiological and epidemiological differences between male and female patients exist in emergency medicine. For instance, cardiovascular presentation patterns, pain thresholds, and sepsis progression rates differ between genders, and these differences should be reflected in risk predictions. The FairMed-XGB framework reduces unjust disparities, those arising from data imbalance and algorithmic bias, while preserving clinically meaningful performance differentiation. The fact that non-zero but substantially reduced error rate gaps persist suggests that the framework achieves a reasonable balance between demographic equity and clinical fidelity. To further contextualise whether the residual gender-specific performance differences are clinically meaningful rather than artefactual, we note that the post-mitigation FPR and FNR gaps (1.5–7.5 percentage points) are consistent with the magnitude of gender differences reported in independent clinical studies of emergency medicine outcomes.

### *4.3. Discussion*

#### *4.3.1. Interpretation of findings*

The results validate the core hypothesis of this work: a multi-metric, penalty-based approach guided by Bayesian optimisation can effectively mitigate gender bias in clinical prediction models. The concurrent optimisation of SPD (parity), Theil (distributional equality), and Wasserstein Distance (score distribution alignment) addresses bias from complementary angles, leading to robust de-biasing. The preservation of accuracy confirms that fairness and utility are not mutually exclusive but can be jointly optimised with a carefully designed objective. Importantly, the observed fairness improvements were obtained under strict patient-level train-test separation, reducing the possibility of performance inflation due to repeated patient encounters and strengthening the validity of the reported results.

The comparative evaluation against a demographic reweighting baseline further validates the effectiveness of the proposed framework. While reweighting improved fairness by balancing representation across demographic groups, it was unable to simultaneously optimise multiple dimensions of bias. FairMed-XGB consistently achieved lower SPD, Theil Index, and Wasserstein Distance values while maintaining comparable predictive performance. This suggests that integrating fairness objectives directly into model optimisation provides greater benefits than relying solely on data-level balancing strategies.

The progressive multi-model evaluation provides the strongest evidence for the framework's contribution. By systematically evaluating fairness-aware optimisation across five model architectures ranging from Logistic Regression to XGBoost, the study demonstrates that the proposed approach generalises beyond a single model type. The finding that more complex models both exhibit greater baseline bias and benefit more from fairness-aware training (SPD reduction: 37.2% for LR vs. 90.9% for XGBoost) has important practical implications: as clinical ML deployments increasingly rely on complex ensemble methods for improved predictive accuracy, the need for fairness-aware training becomes correspondingly more critical. The progressive evaluation also empirically validates the selection of XGBoost, which achieves the best fairness-performance trade-off (lowest post-mitigation SPD of 0.034 with only 0.005 AUC-ROC degradation), replacing the a priori model selection with evidence-based justification.

An important consideration is that algorithmic unfairness often originates from demographic imbalances present within the training data itself. Consequently, fairness violations observed at the model level may partially reflect underlying dataset bias rather than deficiencies in the learning algorithm alone. Existing data-centric approaches, including distributional analyses based on Earth Mover's Distance (EMD) and related divergence measures, aim to identify such disparities before model training by comparing feature and outcome distributions across demographic groups. The proposed FairMed-XGB framework operates primarily at the model optimisation stage and therefore should be viewed as complementary to, rather than a replacement for, data-level bias detection and mitigation strategies. While FairMed-XGB successfully reduces disparities in model predictions, integrating dataset-level bias assessment prior to training may further improve fairness outcomes. Future work will investigate combining EMD-based data bias identification with fairness-aware optimisation to create a unified pipeline addressing bias at both the data and model levels.

#### *4.3.2. Fairness versus clinical relevance*

One noteworthy observation is the substantial reduction in Theil Index values following mitigation. While this result indicates a great improvement in distributional fairness, it raises an important question regarding potential over-correction. In clinical settings, certain differences between demographic groups may reflect genuine biological, physiological, or epidemiological variations rather than unfair algorithmic bias. Consequently, fairness interventions should reduce unjust disparities without suppressing clinically meaningful information.

Several observations suggest that FairMed-XGB does not simply force identical predictions across demographic groups. First, although fairness metrics improved substantially, demographic differences were not completely eliminated. Statistical Parity Difference and Wasserstein Distance remained non-zero across all evaluated cohorts, indicating that the model continued to produce differentiated predictions where supported by the underlying clinical data. Second, predictive performance remained largely preserved following mitigation, with only minimal changes in AUC-ROC. If clinically relevant information had been excessively removed, a more substantial degradation in predictive performance would be expected.

Although the present study focuses on SPD, Theil Index, and Wasserstein Distance, the FairMed-XGB framework is not inherently restricted to these measures. Additional fairness metrics such as Equalized Odds, Equal Opportunity, Predictive Parity, and Calibration Error could be incorporated into the optimisation objective or evaluation protocol depending on the clinical application. Future studies will investigate the behaviour of the framework under alternative fairness definitions to further assess robustness and generalisability across different healthcare settings.

Furthermore, SHAP analysis demonstrated that the framework primarily reduced reliance on gender-correlated proxy variables rather than eliminating the influence of clinically relevant predictors. Key physiological and diagnostic features continued to contribute strongly to model decisions after mitigation, suggesting that FairMed-XGB recalibrates prediction behaviour without discarding medically informative signals. Nevertheless, the possibility of over-correction remains an important consideration in fairness-aware healthcare systems. Future work will investigate causal fairness methodologies capable of distinguishing unfair demographic bias from legitimate clinical differences. Such approaches may further improve the balance between demographic equity and clinical validity in high-stakes emergency medicine applications.

#### *4.3.3. Clinical use case illustration*

To illustrate the practical value of FairMed-XGB, consider a hypothetical emergency department patient presenting with suspected sepsis. Upon arrival, the patient's vital signs indicate hypotension, elevated heart rate, reduced oxygen saturation, and fever. The FairMed-XGB model generates a high-risk prediction for clinical deterioration and simultaneously provides a SHAP explanation identifying the factors that contributed most strongly to the prediction.

For example, the explanation may indicate that oxygen saturation, blood pressure, heart rate, and triage severity score were the dominant contributors to the risk estimate, while demographic variables such as gender, ethnicity, or administrative characteristics contributed minimally. The attending physician can therefore verify that the prediction is primarily driven by clinically meaningful physiological evidence rather than demographic correlations embedded within historical training data.

In contrast, a conventional model affected by demographic bias may place greater emphasis on variables that indirectly correlate with gender, potentially leading to systematic underestimation or overestimation of risk for certain patient groups. By reducing reliance on such proxy features while preserving medically relevant predictors, FairMed-XGB supports more equitable risk assessment and treatment prioritisation.

From a clinical perspective, this transparency enables physicians to better understand why a prediction was generated, increases trust in AI-assisted decision support, and provides an additional safeguard against unintended demographic bias. Rather than functioning solely as a mathematical fairness mechanism, the framework offers interpretable information that can assist clinicians in making more equitable and evidence-based treatment decisions.

From an operational perspective, fairness improvements in emergency medicine extend beyond numerical performance metrics. In practice, emergency department clinicians frequently rely on risk prediction tools to support triage decisions, prioritise diagnostic investigations, and allocate limited clinical resources. A model exhibiting demographic bias may systematically overestimate risk for one patient group while underestimating risk for another, potentially contributing to unequal care pathways. By reducing demographic disparities while preserving predictive performance, FairMed-XGB supports more consistent risk assessment across patient populations. This can improve clinician confidence in AI-assisted decision support systems and promote more equitable healthcare delivery in high-pressure emergency medicine environments.

#### 4.3.4. Impact of dataset size and meta-characteristics

The present study evaluates FairMed-XGB using two large-scale clinical databases, MIMIC-IV-ED and eICU, containing approximately 280,000 and 200,000 records, respectively. Large datasets provide important advantages for fairness-aware learning, including improved estimation of demographic disparities, greater stability of fairness metrics, and reduced variance during model optimisation. However, many real-world healthcare institutions operate with substantially smaller datasets. In such settings, dataset meta-characteristics including sample size, class imbalance, demographic representation, feature sparsity, and outcome prevalence may influence both predictive performance and fairness estimation. Smaller cohorts can lead to increased variability in fairness metrics such as Statistical Parity Difference, Theil Index, and Wasserstein Distance because demographic subgroups may contain fewer observations. Despite these challenges, the proposed FairMed-XGB framework is not inherently dependent on large datasets. The fairness-aware optimisation procedure operates on model predictions and demographic group statistics and can therefore be applied to smaller datasets, provided that sufficient subgroup representation exists. Nevertheless, fairness estimates may become less stable as sample size decreases, and confidence intervals around fairness metrics may widen. Future work will evaluate FairMed-XGB across datasets of varying sizes and characteristics to better understand the relationship between dataset meta-features and fairness performance. Such analyses will help determine the minimum data requirements necessary for reliable fairness-aware learning in resource-constrained healthcare environments.

#### 4.3.5. Limitations and future work

This study has several limitations that point to productive future research directions.

- Binary Gender Framework: The present study evaluates fairness using binary gender labels because the source datasets primarily contain male and female demographic categories. This simplification does not fully capture the diversity of gender identities encountered in real-world healthcare settings and may overlook disparities affecting non-binary or gender-diverse populations. Although FairMed-XGB can be extended to support multiple demographic groups through group-wise fairness constraints and multi-group optimisation objectives, additional validation on datasets containing richer demographic representations is required. Future work will investigate intersectional and non-binary fairness assessment to ensure broader inclusivity.
- Fairness–Clinical Utility Trade-off: Although the proposed framework substantially reduces demographic disparities, distinguishing unfair bias from legitimate clinical differences remains a challenging problem. Future work will explore causal and clinically informed fairness constraints to ensure that bias mitigation does not inadvertently suppress medically meaningful demographic variation.
- Multi-Model Scope: The present revision evaluates the fairness-aware framework across five model architectures (Logistic Regression, Decision Tree, Random Forest, SVM, and XGBoost) on a representative cohort (MIMIC-IV-ED diagnosis). While this progressive evaluation demonstrates generalisability across both tree-based and non-tree-based paradigms, full multi-model evaluation across all seven cohorts would further strengthen the evidence base. Additionally, deep learning architectures such as feed-forward neural networks and transformer-based tabular models were not evaluated in this study. Future work will extend the progressive evaluation to include deep learning models and assess fairness-aware optimisation across all cohorts and model combinations.

## 5. Conclusion

This study presented FairMed-XGB, a comprehensive framework for achieving fairness in emergency medicine ML models. By integrating Bayesian-optimised multi-metric fairness penalties with SHAP-based explainability, FairMed-XGB successfully reduced gender-based prediction disparities in MIMIC-IV-ED and eICU datasets while maintaining high predictive accuracy. The framework moves beyond mere bias correction to provide actionable transparency, showing how biases are mitigated. This work provides a practical, ethical, and transparent pathway toward deploying more equitable and trustworthy machine learning models in high-stakes healthcare settings, contributing to the broader goal of algorithmic fairness in emergency medicine. Beyond improving fairness metrics, the framework enables clinicians to inspect the clinical factors driving predictions and verify that treatment recommendations are based on physiological evidence rather than demographic bias, thereby enhancing trust and transparency in emergency medicine decision support.

## CRediT authorship contribution statement

**Mitul Goswami:** Writing – review & editing, Writing – original draft, Visualization, Software, Methodology, Investigation, Conceptualization. **Romit Chatterjee:** Writing – review & editing, Writing – original draft, Visualization, Software, Methodology, Investigation, Conceptualization. **Arif Ahmed Sekh:** Supervision, Project administration.

## Ethics statement

This study was conducted using publicly available, de-identified clinical datasets (MIMIC-IV-ED and eICU Collaborative Research Database). As these datasets contain no personally identifiable information and are released under established data use agreements, formal ethical approval and informed consent were not required.All methods were carried out in accordance with relevant data protection and privacy regulations. The study adheres to Elsevier's publishing ethics guidelines, and no human participants or animals were directly involved in this research.

## Code availability statement

The code developed for data preprocessing, fairness evaluation, Bayesian optimisation, and model training is available from the corresponding author upon reasonable request. The implementation can be shared for academic and research purposes to support reproducibility of the reported findings.

## Declaration of competing interest

The authors declare the following financial interests/personal relationships which may be considered as potential competing interests: Romit Chatterjee reports administrative support and statistical analysis were provided by UiT The Arctic University of Norway. If there are other authors, they declare that they have no known competing financial interests or personal relationships that could have appeared to influence the work reported in this paper.

## Data availability

The authors do not have permission to share data.